# 3-D Scene Graph: A Sparse and Semantic Representation of Physical Environments for Intelligent Agents

Ue-Hwan Kim, Jin-Man Park, Taek-Jin Song, and Jong-Hwan Kim, *Fellow, IEEE*

*Abstract*—Intelligent agents gather information and perceive semantics within the environments before taking on given tasks. The agents store the collected information in the form of environment models that compactly represent the surrounding environments. The agents, however, can only conduct limited tasks without an efficient and effective environment model. Thus, such an environment model takes a crucial role for the autonomy systems of intelligent agents. We claim the following characteristics for a versatile environment model: accuracy, applicability, usability, and scalability. Although a number of researchers have attempted to develop such models that represent environments precisely to a certain degree, they lack broad applicability, intuitive usability, and satisfactory scalability. To tackle these limitations, we propose 3-D scene graph as an environment model and the 3-D scene graph construction framework. The concise and widely used graph structure readily guarantees usability as well as scalability for 3-D scene graph. We demonstrate the accuracy and applicability of the 3-D scene graph by exhibiting the deployment of the 3-D scene graph in practical applications. Moreover, we verify the performance of the proposed 3-D scene graph and the framework by conducting a series of comprehensive experiments under various conditions.

*Index Terms*—3-D scene graph, environment model, intelligent agent, scene graph, scene understanding.

## I. INTRODUCTION

THE CAPABILITY of understanding the surrounding environments is one of the key factors for intelligent agents to successfully complete given tasks [1]. Without the capability, the agents can only perform simple and limited tasks. For versatile performance, the agents have to perceive not only the physical attributes of the environments but also the semantic information inherent in the environments. In the process of observing the environments and storing up the collected information, the agents construct environment models, which compactly represent the surrounding spaces [2]. Such models include dense maps generated by SLAM [3] and descriptions of scenes [4] produced by computer vision and natural language processing (NLP) algorithms. Environment models let the agents plan how to perform given tasks and offer grounds for inference and reasoning. Thus, an effective environment model for intelligent agents is of great importance.

We claim the following properties as an effective environment model for intelligent agents.
1) *Accuracy:* The model should delineate environments precisely and provide intelligent agents with correct information.
2) *Applicability:* Intelligent agents should be able to utilize the model in performing various types of tasks rather than just one specific task.
3) *Usability:* The model should not require a complicated procedure for usage, but provide intuitive user interface for application.
4) *Scalability:* The model should be able to depict both large- and small-scale environments and increment the coverage step by step.

The claimed properties of environment models enhance the autonomy of intelligent agents. Intelligent agents could collect correct information regarding the environments in the process of constructing the models (accuracy), utilize the models to complete diverse tasks staying in the environments (applicability) using the models with the ease (usability), and expand the knowledge of the environments incrementally (scalability).

In this paper, we define 3-D scene graph, which represents the physical environments in a sparse and semantic way, and propose the 3-D scene graph construction framework. The proposed 3-D scene graph describes the environments compactly by abstracting the environments as graphs, where nodes depict the objects and edges characterize the relations between the pairs of objects. As the proposed 3-D scene graph illustrates the environments in a sparse manner, the graph can cover up an extensive range of physical spaces, which guarantees the scalability. Even when agents have to deal with a broad range of environments or encounter new environments in the middle, 3-D scene graph offers a quick way for accessing and updating the environment models. Furthermore, the concise structure of the 3-D scene graph ensures intuitive usability. The graph structure is straightforward, since 3-D scene graph

Manuscript received October 12, 2018; revised May 20, 2019; accepted July 22, 2019. This work was supported by the Institute for Information and Communications Technology Promotion (IITP) Grant funded by the Korean government (MSIT) (No. 2016-0-00563, Research on Adaptive Machine Learning Technology Development for Intelligent Autonomous Digital Companion). This paper was recommended by Associate Editor P. De Meo. *(Corresponding author: Jong-Hwan Kim.)*

The authors are with the School of Electrical Engineering, Korea Advanced Institute of Science and Technology, Daejeon 34141, South Korea (e-mail: uhkim@rit.kaist.ac.kr; jmpark@rit.kaist.ac.kr; tjsong@rit.kaist.ac.kr; johkim@rit.kaist.ac.kr).

This paper has supplementary downloadable material available at http://ieeexplore.ieee.org, provided by the author.

Color versions of one or more of the figures in this paper are available online at http://ieeexplore.ieee.org.

Digital Object Identifier 10.1109/TCYB.2019.2931042

follows the convention of common graph structures and graph structures are already familiar to majority of the researchers due to its wide use. Next, we verify the applicability of the 3-D scene graph by demonstrating two major applications: 1) visual question and answering (VQA) and 2) task planning. The two applications are under active research in computer vision [5], NLP [6], and robotics societies [7].

The proposed 3-D scene graph construction framework extracts relevant semantics within environments, such as object categories and relations between objects as well as physical attributes, such as 3-D positions and major colors in the process of generating 3-D scene graphs for the given environments. The framework receives a sequence of observations in the form of RGB-D image frames. For robust performance, the framework filters out unstable observations (i.e., blurry images) using the proposed adaptive blurry image rejection (ABIR) algorithm. Then, the framework factors out keyframe groups to avoid redundant processing of the same information. Keyframe groups contain reasonably overlapping frames. Next, the framework extracts semantics and physical attributes within the environments through the recognition modules. During the recognition processes, spurious detections get rejected. Finally, the gathered information gets fused into 3-D scene graph and the graph gets updated upon new observations.

The main contributions of this paper are as follows.
1) We define the concept of the 3-D scene graph which represents the environments in an accurate, applicable, usable, and scalable way.
2) We design the 3-D scene graph construction framework which generates 3-D scene graphs for environments upon receiving a sequence of observations.
3) We provide two application examples of the 3-D scene graph: a) VQA and b) task planning.
4) We conduct a series of thorough experiments and analyze the experiments both quantitatively and qualitatively to verify the performance of the 3-D scene graph.
5) We make the source code of the algorithms presented in this paper public[1] to contribute to the research society and the development of the field.

The remainder of this paper is organized as follows. Section II introduces the related works with associated results and issues. Section III presents an overview of the 3-D scene graph construction framework. Sections IV and V describe each module of the proposed 3-D scene graph construction framework. Section VI provides two major applications of the 3-D scene graph. Section VII illustrates the experimental settings and the analysis of the results. Discussion and concluding remarks follow in Sections VIII and IX, respectively.

## II. Related Works

We overview the previous research outcomes relevant to our 3-D scene graph and point out the differences between the previous works and 3-D scene graph in this section.

[1]https://github.com/Uehwan/3-D-Scene-Graph

### A. Environment Representation

A number of studies have attempted to digitize and store environmental information obtained from various sensors. Although there exist multiple options for sensor selection, we focus on visual sensors in this paper. The studies fall into two categories: 1) raw and dense representations and 2) abstractive and descriptive representations. First of all, the raw and dense representations aim to represent the environments as they are with minimum distortion. They take two steps in general. First, the algorithms gather information for model building using one of the SLAM algorithms [3], [8], and then construct environment models through point clouds [9], grid-based models [10], and probabilistic occupancy maps [11]. The resulting models known as 3-D reconstructions, however, require massive memory space, take much time for processing and lack any semantic information, requiring additional recognition processes to be applicable to reasoning-enabled high-level AI applications.

3-D semantic segmentation, which labels every point in a point cloud, supplements the raw and dense representations with semantic information. The area is under active research and empowered by recent advances in deep learning. One recent work has demonstrated a real-time 3-D semantic segmentation by combining the latest SLAM [3] with a CNN-based 2-D semantic segmentation algorithm [12]. In addition, methods directly segmenting 3-D point clouds semantically have been presented [13]. These methods can provide the locations and classes of objects in given environments. Nevertheless, the relationships between objects are missing in these methods and these methods still require much memory and processing time.

The abstractive and descriptive representations, on the other hand, aim to summarize the characteristics of the environments by remaining only the relevant information. Algorithms for the abstractive and descriptive representations include image captioning, which describes given scenes in natural language [4]. Image captioning algorithms select the pairs of objects from the given images, generate the descriptions for the pairs, and repeat the previous processes multiple times. Therefore, the relations between the objects are naturally extracted in the process of generating descriptions. However, the coverage of the spaces image captioning methods can express is confined to the field of view (FoV) of the cameras, since the methods utilize only one image at a time. In addition, the output of image captioning is in the form of natural language, which makes it difficult for other AI applications to utilize the information as natural language presents the information in an unordered way. Video captioning [14] could compensate for the limited coverage of image captioning, but the current focus of the field is on the dynamics of objects or scenes, discarding considerable amount of information retained in the sequence. Moreover, the output format of video captioning does not vary from that of image captioning.

### B. Scene Graph

A scene graph represents an image with a graph structure, where the nodes correspond to each object instance in the image and the edges depict the pairwise relationships between the objects (see Fig. 1). Compared to the previous text-based representations of visual scenes [4], the scene graph representation offers much contextual information with respect to relative geometry and semantics. The researches related to scene graph can be categorized into two groups: 1) scene graph generation and 2) application of scene graph. The scene graph generation algorithms focus on producing accurate scene graphs given visual scenes. Current state-of-the-art methods utilize the iterative message parsing [15], the multitask formulation technique [16] for boosting the performance, or the subgraph formulation [17].

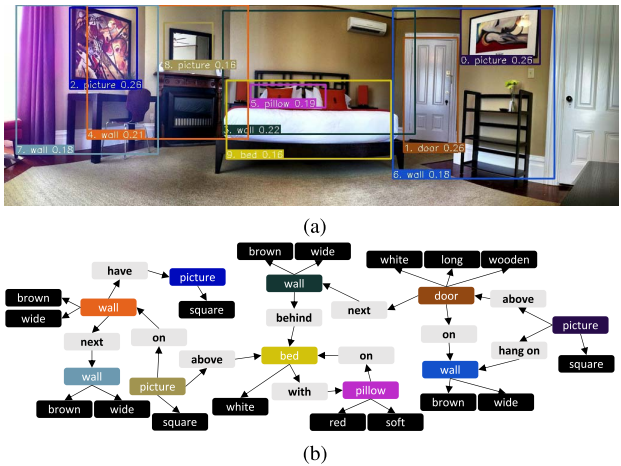

Fig. 1. Example of 2-D scene graph. The scene graph abstracts the given scene and represents the relations between objects (gray nodes) as well as object attributes (black nodes). (a) Input scene with object recognition results. (b) Corresponding scene graph.

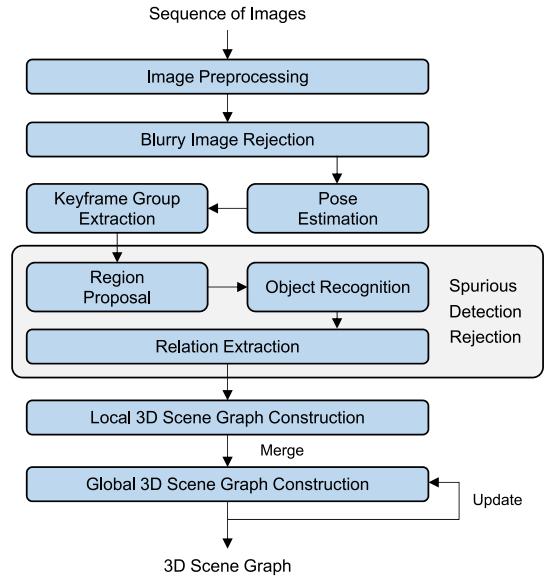

Fig. 2. Overall architecture of the proposed 3-D scene graph construction framework. The framework receives RGB-D image frames and then processes the input to generate 3-D scene graph.

For the second point, a wide range of visual tasks, such as semantic image retrieval [18], scene synthesis [19], and visual question answering [20] employs the scene graph representations and report enhanced performance. As the end-to-end approaches often exploit the statistics of the training dataset rather than truly understanding the visual scenes, the scene graph representation facilitates the algorithms to actually extract relevant features and understand the semantics. The current research on scene graphs, however, concentrates on 2-D static scenes, which cannot provide physical attributes such as 3-D positions, and deal with only one image, which limits the spatial coverage. Our proposed 3-D scene graph expands 2-D scene graphs into 3-D spaces. This expansion enables intelligent agents to understand the surrounding environments in a more tangible way and to perform tasks in a more stable and precise manner.

## III. OVERALL SYSTEM ARCHITECTURE

Fig. 2 shows the overall system architecture of the proposed 3-D scene graph construction framework. The proposed framework takes in streams of images and preprocesses the input images. In the preprocessing step, the noise within the images gets removed and appropriate scaling and cropping are applied. Then, the preprocessed images enter the blurry image rejection module. Blurry images get rejected to guarantee the stable performance of later modules. Then, the pose estimation module extracts the relative poses between nearby frames using visual odometry or SLAM. The poses estimated in this module are utilized for extracting the keyframe groups and physical features of objects in the following modules. In this framework, we use ElasticFusion [3] and BundleFusion [8] for estimating the poses. Next, the keyframe group extraction (KGE) module categorizes each frame as key, anchor, or garbage frames for boosting efficiency. After KGE, the framework processes only the reasonably overlapping frames.

The extracted keyframe groups go through the region proposal and object recognition modules. The region proposal module detects the plausible regions of object instances and the object recognition module classifies the objects within the regions. Specifically, we use Faster-RCNN [21] to recognize object instances (region proposal) and object categories (object recognition). The object recognition module outputs a list of category candidates with confidence scores for an object. After the object instances with the corresponding classes are extracted, the relation extraction module identifies the pairwise relations between object instances. The relations include action (e.g., jumping over), spatial (behind), and prepositional (e.g., with) relations. Factorizable Net (F-Net) [17] is incorporated in the proposed framework for the relation extraction process.

Running the recognition modules on the stream of images inevitably generates spurious detections despite postprocessing procedures followed by the recognition modules. The spurious detection rejection (SDR) module removes both the spurious and duplicate detections using 3-D position information, semantics from word2vec [22], and rule-bases. The local 3-D scene graph construction module receives the processed detection results and constructs a local 3-D scene graph for one input frame. This local 3-D scene graph covers only a short range of physical spaces and the first local 3-D scene graph becomes the initial global 3-D scene graph. In the global 3-D scene graph construction module, the initial global 3-D scene graph gets updated and expanded upon receiving local 3-D scene graphs generated from the following image frames.

## IV. DATA PROCESSING

In this section, we describe the major data processing modules of the proposed 3-D scene graph construction framework. The three modules proposed in this paper allow the efficient and effective generation of the 3-D scene graphs.

### A. Adaptive Blurry Image Rejection

For stable performance, the object recognition and relation extraction modules require clean and motionless images as input. Blurry input images deteriorate the performance, since the shape, size, and even color of an object appear different in the blurry images. The input image sequences for the proposed 3-D scene graph construction framework could contain such blurry images due to abrupt camera motions. To prevent blurry images and to guarantee the robust performance of the recognition modules, we adopt the variance of Laplacian defined as

$$V = \frac{1}{WH}\sum_{x=1}^{W}\sum_{y=1}^{H} L(x,y)^2 - \left\{\frac{1}{WH}\sum_{x=1}^{W}\sum_{y=1}^{H} L(x,y)\right\}^2 \quad (1)$$

where $W$ and $H$ denote the image width and height, respectively, and $L(x,y) = (\partial^2 I/\partial x^2) + (\partial^2 I/\partial y^2)$ is the Laplacian operator. The variance of Laplacian measures the intensity variations across pixels in an image. By filtering images with the variance of Laplacian less than a threshold, blurry images get removed. However, a fixed threshold value could detect nonblurry images with low texture as blurry frames, since the intensities of low texture images do not vary significantly across pixels.

To cope with the varying texture problem, we propose the ABIR algorithm. First, the proposed algorithm evaluates the exponential moving average (EMA), $S_t$, over the Laplacian variances

$$S_t = \begin{cases} V_t & t=1 \\ \alpha \cdot S_{t-1} + (1-\alpha) \cdot V_t & t>1 \end{cases} \quad (2)$$

where $t$ is the time step, $V_t$ is the variance of Laplacian at $t$, and $\alpha$ is a constant smoothing factor in the interval [0, 1]. A lower $\alpha$ reduces the effect of older observations faster. At the initial phase, EMA does not follow the observed values since only a small number of observations are available at first. This phenomenon is known as a bias and we correct the bias as follows:

$$S'_t = \frac{S_t}{1-\alpha^t}. \quad (3)$$

Then, we modify the bias-corrected average value, $S'_t$, to generate the adaptive threshold, $t_{\text{blurry}}$, as follows:

$$t_{\text{blurry}} = g \cdot \ln(1+S'_t) + b \quad (4)$$

where $g$ and $b$ represent the gain and offset, respectively.

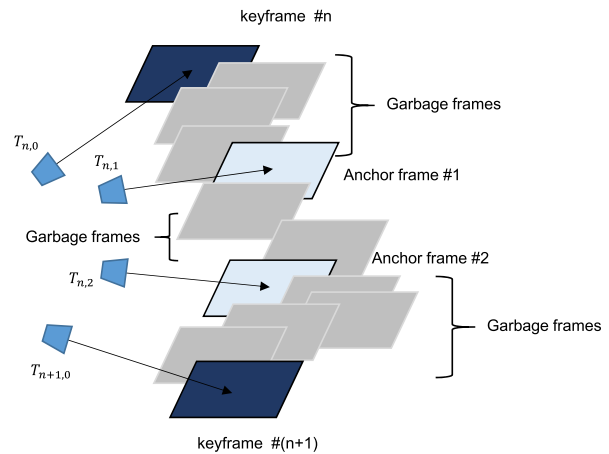

Fig. 3. Process of KGE. The input image frames are classified into three classes: key, anchor, and garbage frames. Through the process, redundancies in the image frames are resolved.

### B. Keyframe Group Extraction

The KGE module 1) receives the sequence of preprocessed image frames; 2) filters out unnecessary frames by categorizing the input frames into three classes; and 3) forms the keyframe groups. The three classes of the input frames are as follows.
1) *Keyframe:* A keyframe works as a reference. The keyframe determines the first anchor frame and the coverage of the keyframe group.
2) *Anchor Frame:* An anchor frame is either active or inactive. Only the latest anchor frame is active and the active anchor frame determines the next anchor frame.
3) *Garbage Frame:* Frames other than the keyframes and the anchor frames are classified as garbage frames and discarded due to redundancy.

Fig. 3 and Algorithm 1 depict the process of the KGE module. First of all, the module sets the first nonblurry incoming frame as the first keyframe. The rest of the frames go through the classification process. Each incoming frame is compared with both the current keyframe and the active anchor frame. The frame with less than $t_{\text{anchor}}$ % of overlap with the active anchor frame is kept as the next anchor frame. For the extraction of the first anchor frame, the input frames are compared with the keyframe. With the detection of a new anchor frame, the current active anchor frame becomes inactive and the new anchor frame turns into active. If an incoming frame overlaps less than $t_{\text{keyframe}}$ % with the keyframe, the incoming frame becomes a new keyframe and the previous keyframe and the anchor frames detected up to the point form a keyframe group. We set $t_{\text{anchor}}$ larger than $t_{\text{keyframe}}$.

The overlap between two frames is calculated by projecting one frame into another frame's coordinate as follows:

$$\text{overlap} = \frac{1}{W \cdot H}\sum_{x}\sum_{y} \vec{\mathbf{1}}_{I_{W,H}}(p'(i,j)) \quad (5)$$

where $\vec{\mathbf{1}}_A(\cdot)$ is an indicator function for a set $A$, $I_{W,H} = \{(x,y) | 0 \le x < W, 0 \le y < H, (x,y) \in \mathbb{Z}^2\}$, and $p'(i,j)$ is a projection function from source image frame to target image

**Algorithm 1** Keyframe Group Extraction

**Input**: Sequence of image frames
**Output**: List of keyframe groups
1: keyframe_groups = []
2: keyframe, anchor = [get_next_frame()] * 2
3: curr_keyframe_group = [keyframe]
4: **while** curr_frame = get_next_frame() **do**
5:     # compute overlap by Eq. (5)
6:     overlap_key = overlap(keyframe, curr_frame)
7:     overlap_anchor = overlap(anchor, curr_frame)
8:     # update keyframe and anchor
9:     **if** overlap_key $< t_{key}$ **then**
10:         keyframe_groups.append(curr_keyframe_group)
11:         keyframe, anchor = [curr_frame] * 2
12:         curr_keyframe_group = [keyframe]
13:         continue
14:     **end if**
15:     **if** overlap_anchor $< t_{anchor}$ **then**
16:         anchor_frame = curr_frame
17:         curr_keyframe_group.append(anchor)
18:     **end if**
19: **end while**

frame, which is defined as

$$p' = K \cdot T_{i,j} \cdot D(p) \cdot K^{-1} \cdot p \quad (6)$$

where $K$ is the camera intrinsic, $T_{i,j}$ is the relative pose between the $i$th and $j$th frames, $D$ is the depth measure of point $p$, and $p$ is the original point before projection.

By having both keyframe and anchor frame as two references, the proposed KGE can effectively handle the redundant information inherent within consecutive image sequences. In addition, the proposed framework does not explode, but efficiently deals with the redundancies even when the camera captures the input frames staying around at the same point for a long time. The time complexity of Algorithm 1 is $O(NWH)$, where $N$ denotes the number of input image frames. The single while loop in Algorithm 1 makes the total time complexity $O(N \cdot \text{overlap})$. The time complexity of the overlap function is $O(WH)$ since the function iterates through both the horizontal and vertical directions as shown in (5) and each projection takes constant time as shown in (6).

### C. Spurious Detection Rejection

Processing streams of images unavoidably produces the erroneous results, since the recognition modules do not achieve the perfect performance and image frames captured with the sensing devices are corrupted with noise. The purpose of SDR lies in removing both the spurious and repeated detections by distilling human knowledge and priors from other knowledge bases. SDR works over multiple modules: region proposal, object recognition, and relation extraction modules. First, SDR removes redundant object regions proposed by the region proposal module. It utilizes the popular nonmaximum suppression (NMS) [23] to achieve the goal. For the objection recognition module, the SDR module removes the irrelevant objects.

Moreover, SDR deletes the predefined irrelevant objects, such as road, sky, building, and moving objects. These objects cannot exist in the setting of the 3-D scene graph.

For the relation extraction module, the SDR module inspects the detected relations between the pairs of objects from multiple frames in one keyframe group and the most-occurred relations remain in the graph. If a tie occurs, all the top-ranked relations get added to the graph. Then, SDR applies a relation dictionary as a prior. The relation dictionary stores the statistics of relations for the pairs of objects. We construct the relation dictionary using the visual genome dataset [24] in advance. We collect the statistics of the relations between the pairs of objects and store the pixel distance information ($d_{\text{pixel}}$) as the Gaussian distributions.

To apply the prior, SDR gathers the related object pairs, searches through the relation dictionary, and calculates the probabilities of the detected relations. The relations with the probabilities lower than a threshold value are deleted from the graph. In calculating the probabilities, we utilize the pixel distances between the objects to filter out relations between the pairs of objects with the distances the relations cannot make sense. The probability of a relation, $Pr(r|d_{\text{pixel}})$, takes both the pixel distance and frequency into account as follows:

$$Pr(r|d_{\text{pixel}}) = Pr_{dict}(r|d_{\text{pixel}}) \cdot \phi_{\mu,\sigma^2}(d_{\text{pixel}})/\phi_{\mu,\sigma^2}(\mu) \quad (7)$$

where $Pr_{\text{dict}}(r|d_{\text{pixel}})$ is the statistics from the visual genome dataset and $\phi_{\mu,\sigma^2}(k) = (1/\sqrt{2\pi\sigma^2})\exp(-[1/2]([k-\mu]/\sigma)^2)$ is the Gaussian function. We approximate the probability of distance between the points by a normalized probability density function.

## V. 3-D Scene Graph Construction

In this section, we define the proposed 3-D scene graph with its properties. Next, we detail the 3-D scene graph construction algorithm and the merge and update algorithm for the global 3-D scene graph.

### A. Graph Representation

A 3-D scene graph $G$ defines a pair of sets $G = (V, E)$, where $V$ and $E$ denote the set of vertices and edges, respectively. On the one hand, the vertices represent the objects present in the environments. Each vertex contains the information regarding the object it represents. The information contained in each vertex is as follows.
1) *Identification Number (ID):* A unique number assigned to each object in the environment.
2) *Semantic Label:* Category of the object classified by the object recognition module.
3) *Physical Attributes:* Physical characteristics, such as size (height and width), major colors, or position relative to the first keyframe.
4) *Visual Feature:* A thumbnail, color histogram, or extracted visual features with descriptors.

On the other hand, the following lists the types of relations an edge could stand for.
1) *Actions:* Behaviors shown by one object toward another one (e.g., feeding).

2) *Spatial Relation:* Spacious relations, such as distance and relative position (e.g., in front of).
3) *Description:* States of one object related by another object (e.g., wear).
4) *Preposition:* Semantic relations that are expressed by prepositions (e.g., with).
5) *Comparison:* Relative attributes of one object compared to another one (e.g., smaller).

As one object is subjective and the other is objective given a pair of objects, the edges in 3-D scene graphs are directed (the subjective objects point at the objective objects). Therefore, 3-D scene graphs are directed graphs (or digraphs).

### B. Local Graph Construction

Up to the relation extraction module, the proposed 3-D scene graph construction framework extracts a part of object attributes and the relations between objects, which could form a 2-D version of scene graph for the input frame. In the local 3-D scene graph construction module, the 2-D attributes are translated into 3-D attributes and additional attributes are extracted. The constructed local 3-D scene graph for one image frame gets merged and updated into the global 3-D scene graph in the following module.

The ID of an object is temporarily assigned and later module determines the ID in the respect of the global 3-D scene graph. Furthermore, the previous modules readily provide the semantic label candidates for the object with the corresponding scores. We keep the top-$k$ labels and the relevant scores for the same node detection afterward. The 3-D position of the object, on the other hand, is expressed as a Gaussian distribution, because using only one center point for the position of the object is prone to measurement error. We carve out the center rectangle from the object bounding box given by the region proposal module, after dividing the object region into $5 \times 5$ subregions. Then, the 3-D position of each point in the center rectangle relative to the first keyframe, $p''$, is evaluated by

$$p'' = T_{i,o} \cdot D(p) \cdot K^{-1} \cdot p \tag{8}$$

where $i$ and $o$ represent the indices of the current frame and the first keyframe, respectively. Then, we evaluate the mean ($\mu$) and variance ($\sigma^2$) for the Gaussian distribution of the 3-D position $\sim \mathcal{N}(\mu, \sigma^2)$. In the process, we assume independent identically distributed for each dimension, $x$, $y$, and $z$, and keep the number of points used for the evaluation.

Next, we calculate the color histogram for the object, $h_{H,S,V}$, as follows:

$$h_{H,S,V} = N \cdot Pr(H = h, S = s, V = v) \tag{9}$$

where $(H, S, V)$ represents the three axes of color space and $N$ is the number of pixels. It is reported that $(H, S, V)$ space shows the superior performance over $(R, G, B)$ space in general [25]. We digitize each axis of the color space into $c$-bins, making the size of the histogram $c^3$. Finally, the region inside the bounding box becomes the thumbnail of the object.

The attributes extracted in this step get updated and modified by later modules which collect the information of objects from multiple frames and then make the final decisions.

### C. Graph Merge and Update

The graph merge and update process merges individual 3-D scene graphs for single image frames into one global 3-D scene graph and updates the nodes and edges of the global 3-D scene graph accordingly. As the camera view and position vary, the recognition module would extract different information from the same objects. The followings compensate for such variations and construct the global 3-D scene graph for the entire environment.

*1) Same Node Detection:* Without same node detection, 3-D scene graph would explode with same nodes added numerous times and multiple observations of the same objects cannot be integrated effectively. We utilize the following features for the same node detection: object label, 3-D position, and color histogram. From these features, we evaluate the similarity scores as follows.

First, we define the label similarity, $s_{\text{label}}$, as

$$s_{\text{label}} = \begin{cases} |C_o \cap C_c| \cdot \text{score} & |C_o \cap C_c| > 0 \\ \{1 - d_{wv}(f_{wv}(l_o), f_{wv}(l_c))\} \cdot \text{score} & \text{otherwise} \end{cases} \tag{10}$$

where the subscripts $o$ and $c$ refer to the original node in the global 3-D scene graph and candidate node in the current frame, respectively, $C_o$ and $C_c$ contain the top-$k$ labels for the objects, and $l_o$ and $l_c$ are the labels of the objects with the maximum scores. The number of common elements complements the score when the two label sets share the same elements, while the score gets penalized by the distance between word vectors, otherwise. The score for the label similarity is evaluated by

$$\text{score} = \max_{i \in \{o,c\}} \{f_{s_i}(l) : l \in C_i\} \tag{11}$$

where the score function, $f_{s_i}$, returns a score for the given label $l$ in the candidate set, $C_i$.

Second, the position similarity, $s_{\text{position}}$, is evaluated as

$$s_{\text{position}} = \prod_{j \in \{x,y,z\}} I^j(\mu_c^j) \tag{12}$$

where $\mu_c$ is the mean position of the object in the candidate node and $I^j$ calculates the position similarity of $x$, $y$, and $z$ as follows:

$$I^j = \begin{cases} 1 & |\mu_c^j - \mu_o^j| < \sigma_o^j \\ \dfrac{1 - \phi(Z_{\mu_c^j}) + \phi(-Z_{\mu_c^j})}{1 - \phi(Z_{\sigma_o^j}) + \phi(-Z_{\sigma_o^j})} & \text{otherwise} \end{cases} \tag{13}$$

where $Z$ is the $z$-score of the normal distribution and $\phi(\cdot)$ returns the area of the standard normal distribution. If the candidate object positions within the $\sigma_o$ boundary of the object in the global 3-D scene graph, no penalty occurs. Otherwise, the score becomes inversely proportional to the distance.

Third, we calculate the following color similarity, $d_h(h^i, h^j)$, using the intersection of histograms:

$$d_h(h^i, h^j) = \frac{\sum_X \sum_Y \sum_Z \min(h^i(x,y,z), h^j(x,y,z))}{\min(|h^i|, |h^j|)} \tag{14}$$

where $h^i$ and $h^j$ are a pair of histograms for comparison, $X$, $Y$, and $Z$ are the axes of 3-D space, and $|\cdot|$ gives

the magnitude of a histogram. Among various options for histogram distance measure, the intersection of histogram guarantees efficient computation as well as effective comparison for color histograms [25]. The color similarity becomes $s_{\text{color}} = 1 - d_h(h^o_{H,S,V}, h^c_{H,S,V})$.

Finally, the confidence score for the same node, $s_{\text{total}}$, is the weighted combination of the three similarities

$$s_{\text{total}} = \sum_{i \in F} w_i \cdot s_i \qquad (15)$$

where $F = \{\text{label, color, position}\}$ contains the features. A pair of nodes with the similarity score higher than a threshold is detected as the same node.

*2) Merge and Update:* The 3-D scene graph constructed by the first keyframe forms the initial global 3-D scene graph for the environment. Upon receiving image frames, local 3-D scene graphs are generated and merged to the global 3-D scene graph. The merge process compares the nodes in the newly constructed scene graph with the nodes in the global 3-D scene graph and detects same nodes. Nodes not included in the global 3-D scene graph get added to the graph with the corresponding edges and nodes detected as same nodes update the nodes in the global graph. The update process proceeds as follows. The label set for the same node selects the top-$k$ labels with the maximum scores from $C_o \cup C_c$. The 3-D position now considers newly sampled points and recalculates the mean and variance. The color histogram gets combined with the incoming color histogram. The number of points kept in the node becomes the number of the original points plus the number of new points. The thumbnail gets replaced with the incoming one if the label of the maximum score comes from the incoming scene graph.

## VI. Application: VQA and Task Planning

The proposed 3-D scene graph allows deeper understanding of the environments for intelligent agents. Thus, the agents can perform various tasks in a versatile manner. Such tasks include VQA, task planning, 3-D space captioning, 3-D environment model generation, and place recognition. We illustrate two of the major applications of the 3-D scene graph in this section.

### A. Visual Question and Answering

It is possible to assess how well intelligent agents understood a given entity, such as image or text by asking questions and evaluating the answers replied. In this paper, we adopt a similar question and answering (QA) approach both to demonstrate the performance of the 3-D scene graph in environment understanding for intelligent agents and to provide one of the major applications of the 3-D scene graph.

Contrast to ongoing research on the QA approach where QA pairs along with the entity to understand are provided as the training datasets, such QA datasets are not prepared for 3-D scene graph as the area is in the commencing stage. Therefore, we constrain the types of questions as follows.

1) *Object Counting:* Either simple (e.g., how many cups are in the environment?) or hierarchical (e.g., how many pieces of cutlery are there?).

2) *Counting With Attributes:* Number of objects with specific attributes, such as size, visual feature, and location (e.g., how many red chairs are in the environment?).
3) *Counting With Relations:* Number of objects distinctively related to a specific object (e.g., how many objects are on the shelf?).
4) *Multimodal VQA:* An answer by providing a thumbnail of an object (e.g., show me the biggest bowl in the environment).

The above-listed questions can be easily converted to a query-form (machine readable form) even when asked in a free-form and open-ended manner by employing a few NLP algorithms. The intelligent agent could answer the questions by searching the asked entity through the constructed 3-D scene graph. A number of graph search algorithms are available off the shelf, including depth first search and breadth first search.

### B. Task Planning

Robots plan how to perform given tasks before actually conducting the tasks, which is called task planning. In the process of task planning, robots generate the sequences of primitive actions to achieve the goals using the collected environment information. The environment information includes the positions, states, and attributes of objects in the environment. Task planning assumes the environment information is already given before it starts to generate the action plans. Therefore, the environment information, the purpose of the 3-D scene graph, takes a central role in task planning.

We demonstrate the effectiveness of the 3-D scene graph in task planning using fast-forward (FF) planner [26], one of the representative examples of task planner. FF planner requires two types of descriptions: 1) problem description and 2) domain description. The problem description contains the information regarding the categories and states of objects and the goal definition. The domain description describes the primitive actions robots can take. The descriptions are written in planning domain definition language (PDDL) [27]. The 3-D scene graphs can be directly turned into the problem description format, as it stores the information FF planner needs. A few rule-bases can transform 3-D scene graph into problem descriptions.

## VII. Experiment

In this section, we first focus on the verification of accuracy, since the widely used and well-known graph structure already guarantees both usability and scalability. Next, the applicability of the 3-D scene graph is verified.

### A. Performance Verification

*1) Dataset:* We selected a few sequences from the ScanNet dataset [28]. ScanNet consists of 1513 sequences, which were collected using one type of RGBD-sensor. The resolutions of the image frames are $1269 \times 968$ (color) and $640 \times 480$ (depth) with 30-Hz frame rate and all the image and depth frames are calibrated. The imaging parameters for rgb and depth sensors are provided as well. Classes of imaging environments include bedroom, classroom, office, apartment, etc. We chose one challenging sequence after filtering out sequences with too few objects, narrow coverages, or multiple blurry images. The length of the sequence is 5578 with the sequence number 0 of a living room.

TABLE I
LIST OF ALGORITHMS FOR COMPARATIVE STUDY

| Methods | Modules |
|---|---|
| base | region proposal, object recognition, relation extraction |
| 2D-basic | base + same triple pair detection |
| 2D-efficient | 2D-basic + ABIR + KGE |
| 3D-basic | base + 3D attributes + same node detection |
| 3D-efficient | 3D-basic + ABIR + KGE |
| 3D-full | 3D-efficient + SDR |

*2) Algorithms:* We compared the quality of the 3-D scene graph with a few baseline methods. We used the extended versions of base scene graph generation algorithm (F-Net) [17] as baselines (see Table I). The first baseline computes 2-D scene graphs for every image frame, then concatenates all the generated graphs after removing the same (subject, relation, object) pairs. The second baseline applies ABIR and KGE to boost efficiency. The third baseline constructs 3-D scene graphs for every input frame and combines the graphs using the proposed same node detection. The fourth baseline is built upon the third baseline, which additionally employs ABIR and KGE for efficiency. The last algorithm refers to the full model, the proposed 3-D scene graph construction framework, which adds SDR to the fourth baseline.

*3) Evaluation Metrics:* We used a human judgment metric to evaluate the accuracy of each method. We recruited six experiment participants (age: 22–35, male/female: 5/1) and gathered six responses for each graph. For each resulting graph, we showed the participants, their corresponding ScanNet sequence and asked to count the number of spurious entities (nodes and edges) and missing entities. We followed the majority voting approach [29] when interpreting the responses, thus only the entities that two thirds of participants responded consistently were considered for the statistics. In addition, we required the participants to rate how well the generated graphs represented the environments on a 7-point Likert scale (overall accuracy). After all, we averaged the results for each method. We measured the runtime of each method to compare the computational efficiency as well.

*4) Implementation Details:* For the implementation of the proposed 3-D scene graph construction framework, we used Python and Pytorch for seamless integration, since the building blocks of the framework had been developed in the same environment. We used Intel Core i9-7980XE CPU@2.60 GHz and Titan XP for the experiment. For ABIR, we set $\alpha$, $g$, and $b$ as 0.9, 30, and 25, respectively. We tuned the values before applying the algorithm. During the process of calculating overlaps between the frames for KGE, we sampled 1000 points after projection from source to target frame to reduce the amount of computation. Inspecting all the projected points required much computation, slowing down the entire process. In addition, SDR rejects 68 predefined object classes among 400 classes the recognition module can classify. To reject spurious relations, we used 0.5 as the probability threshold. Relations with the probabilities lower than the threshold were rejected. In building color histograms, we divided each axis of the color space into eight bins, resulting in histogram size of 512. For same node detection, we set $w_{\text{label}}$, $w_{\text{color}}$, and $w_{\text{position}}$ as 0.375, 0.25, and 0.375, respectively, and used 0.7 as the threshold. We stored the resulting 3-D scene graphs in the JSON format.

*5) Results and Analysis:* Fig. 4 shows the resulting 2-D and 3-D scene graphs for the experiment sequence with the first keyframe group and Table II reports the quantitative result of the comparative study. The resulting scene graph from the 2D-basic baseline was too crowded by spurious nodes and edges, thus human judges were not able to count the number of spurious or missing entities. Similarly, the resulting 3-D scene graph from the 3D-basic baseline contained multiple spurious nodes. However, 3-D scene graph from the 3D-basic method includes a reasonable number of spurious nodes in contrast to 2D-basic. Comparison between 2D-basic and 3D-basic establishes the superior performance of the same node detection over the simple triple pair detection.

Comparing basic and efficient baselines verifies the performance of ABIR and KGE in boosting the efficiency. The average runtime improves in both 2-D and 3-D cases. The two modules boost the speed of graph construction by avoiding repeated processing of redundant information. The ABIR module, in addition, filters out blurry images, which improves the performance of the recognition modules. Although the ABIR and the KGE modules advance the efficiency, the resulting 2-D and 3-D scene graphs still contain multiple spurious entities.

The proposed framework, the 3D-full model, compactly extracts the objects in the environments and the pairwise relations among the objects. Although a few entities are missing in the resulting 3-D scene graph, SDR rejects most of the irrelevant relations and the quantitative analysis corroborates this. The number of detected objects in the 3-D scene graph from the 3D-full model equals to that of the 3D-efficient. Although the percentages of spurious entities are higher for other baselines than the 3D-full method, human judges assessed the overall accuracy of 3D-full lower than 3D-efficient. We assume that human judges consider the excessiveness better than the missing. The 3D-full method processes the input frames faster than the 3D-efficient method, because the 3D-full method holds less number of entities in the process of constructing 3-D scene graphs. As less number of entities are compared against incoming entities in local 3-D scene graphs, the amount of computation reduces. Human judges were not able to count the number of missing relations for 2D-efficient and 3D-efficient, since the graphs were already crowded by spurious relations.

SDR could be utilized for inferring missing entities. An inference module can figure out missing relations by collecting unpaired objects and calculating the probability of existence of relations between them. In a similar way, missing objects can be inferred by calculating conditional probabilities. We implemented the inference module, but the module did not show noticeable performance improvement. We suspect the knowledge-bases used for priors were not enough to assure the performance. Moreover, the limited performance of the recognition module stems from the inherent feature of the ScanNet dataset whose color images are blurrier than the general image datasets. The average blurriness of the sequence used for the experiment was around 200, whereas the general image datasets provide images with blurriness higher than 1000. The recognition modules trained on the general image datasets would have shown higher performance in the experiment, if they had been trained on blurrier images.

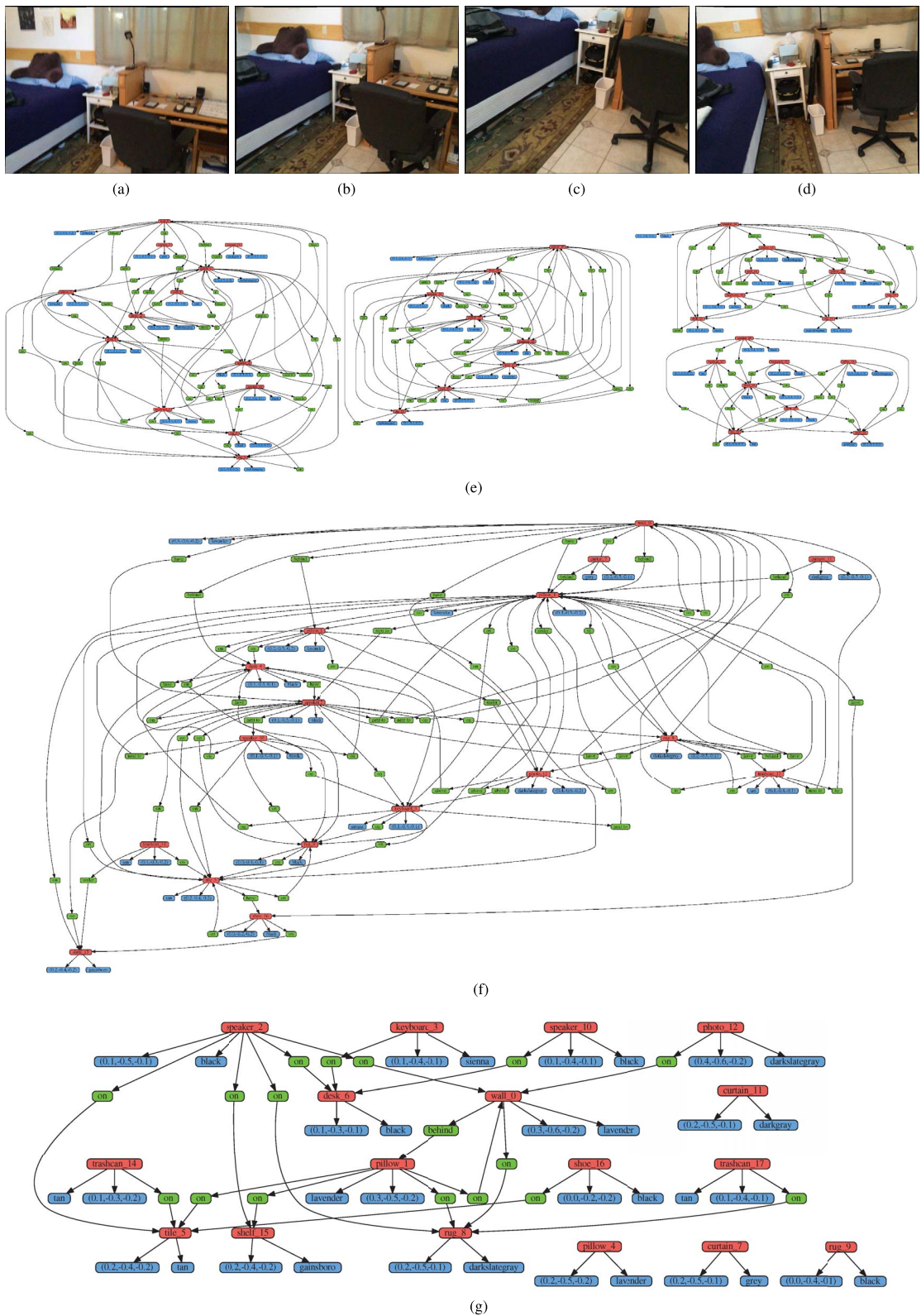

Fig. 4. First keyframe group of the first experiment sequence and the resulting 2-D and 3-D scene graphs. Red, green, and blue denote node, edge, and attributes, respectively. Graphs generated by 2D-basic and 3D-basic are omitted due to over crowded entities. From the basic to the full model, 3-D scene graphs become more efficient, precise, and free of spurious entities. (a) First keyframe. (b) Anchor frame 1. (c) Anchor frame 2. (d) Anchor frame 3. (e) Scene graph from 2D-efficient. (f) 3-D scene graph from 3D-efficient. (g) 3-D scene graph from 3D-full.

TABLE II
RESULTS OF COMPARATIVE STUDY

| Methods | Spurious Entity | | Missing Entity | | Overall Accuracy | Average Runtime |
|---|---|---|---|---|---|---|
| | Node | Edge | Node | Edge | | |
| 2D-basic | $< \infty$ | $< \infty$ | $< \infty$ | $< \infty$ | - | 20.3 (sec/frame) |
| 2D-efficient | 18/37 | 83/128 | 3 | - | 2.67 | 0.84 (sec/frame) |
| 3D-basic | $< \infty$ | $< \infty$ | $< \infty$ | $< \infty$ | - | 3.84 (sec/frame) |
| 3D-efficient | 3/17 | 23/64 | 3 | - | 4.83 | 0.46 (sec/frame) |
| 3D-full | 3/17 | 3/17 | 3 | 2 | 4.17 | 0.38 (sec/frame) |

In summary, the experiment results verified the performance of the proposed 3-D scene graph construction framework. As each module of the proposed framework was added to the base algorithm one by one, the performance improved step by step. The results also established that mere extension of 2-D scene graph generation algorithm to 3-D spaces would not work, but with the proposed 3-D scene graph construction framework.

### B. Applicability Demonstration

*1) Environment:* We constructed a kitchen simulation environment to verify the applicability of the proposed 3-D scene graph. The kitchen simulation environment models the actual kitchen environment, so that the 3-D scene graph from the actual kitchen can be directly used in the simulation. In the simulation environment, human and a robot interact through input devices, such as mouse and keyboard and the robot performs the given tasks. We conducted the experiment with simulated Mybot, which was developed in the Robot Intelligence Technology (RIT) Lab at KAIST. Mybot includes a robotic head connected to the upper body through a three degrees of freedom (DoFs) neck, RGB-D camera, two arms (10 DoFs for each) attached to the upper body, one 2-DoFs trunk, and an omnidirectional wheel-base with a power supply. Mybot is able to perform home chore tasks [30], control gaze [31], and converse autonomously [32]. We utilized previously built modules in the Mybot system to implement the demonstration scenario. We implemented the simulation environment using Webots [33] and ROS [34].

*2) Scenario:* The demonstration scenario consists of four phases. In the first phase, human scans through the actual kitchen and a 3-D scene graph is constructed for being used in the simulation. The generated 3-D scene graph is transferred to Mybot in the simulated environment and Mybot builds a 3-D map of the simulation environment for localization, which is required for the later phases. Next, human asks a couple of questions regarding the environment and Mybot answers those questions. The types of questions human can ask are predefined as discussed before. Third, human commands Mybot to clean up the kitchen (throwing away used cups into the sink) and Mybot generates a problem description using the transferred 3-D scene graph and a sequence of actions for the task using FF planner. Finally, Mybot completes the given task according to the generated task plan.

*3) Results and Analysis:* Fig. 5 shows the demonstration process. In the first step, human scanned through the environment and gathered relevant information to construct a 3-D scene graph. During the process, objects in the environment were detected and the categories, positions, and other attributes of each object were extracted. Mybot in the simulation environment saved transferred 3-D scene graph and built the 3-D map of the simulation environment. Then, Mybot answered questions regarding the environment. Mybot could count the number of objects with specific classes and features and tell how objects were related. After VQA, human commanded Mybot to clean up the table. To accomplish the command, Mybot generated a problem description using the constructed 3-D scene graph. A few rule-bases could transform 3-D scene graph into the problem description in PDDL. Then, Mybot developed a task plan combining the problem description with the predefined domain description. With the task plan, Mybot took cups into the sink one by one.

In short, the demonstration verified the broad applicability of the proposed 3-D scene graph. Intelligent agents can answer questions regarding the environments the agents are situated in utilizing the constructed 3-D scene graphs. 3-D scene graph offers a straight-forward way for intelligent agents to count objects with specific traits and categories. Furthermore, intelligent agents can answer questions asking relations between the pairs of objects as 3-D scene graph characterizes the relations between objects. In addition, the agents can perform given tasks by autonomously formulating problem descriptions in PDDL using 3-D scene graphs. In general, problem descriptions are written by humans. Intelligent agents equipped with 3-D scene graphs can plan tasks with enhanced autonomy.

## VIII. DISCUSSION

In this section, we present a couple of discussion points for future works. First of all, the current setting of the 3-D scene graph does not include moving objects. Although intelligent agents could conduct heaps of tasks in such static environments, real-world settings might contain dynamic objects. One of the dynamic objects that 3-D scene graph should consider with high priority is human, since intelligent agents deployed in real-world entail the issue of interaction with human and security of human. To take dynamic objects into account, research on real-time update of the 3-D scene graph should follow. The real-time update algorithm would handle the same objects with varying positions, changes in relations, and state variations.

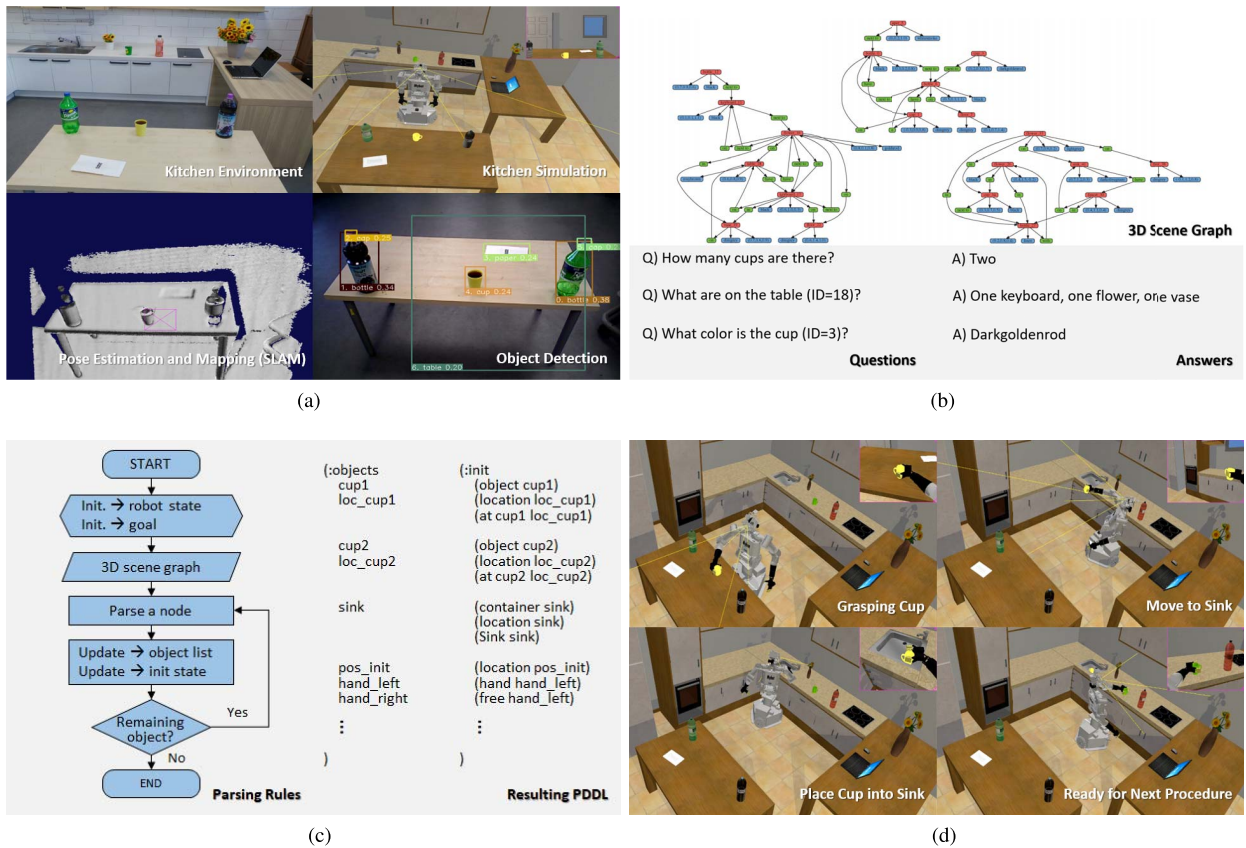

Fig. 5. Applicability demonstration. Human scanned through the kitchen environment and gathered relevant information for constructing a 3-D scene graph. Then, simulated Mybot answered questions regarding the environment using the transferred 3-D scene graph. Upon receiving the command, cleaning up the table, Mybot generated a task plan using the 3-D scene graph. Finally, Mybot performed the given task in the simulated kitchen. (a) Scanning through the environment to construct a 3-D scene graph. (b) VQA with Mybot. (c) Mybot generating problem PDDL from 3-D scene graph. (d) Mybot performing the given task.

In the second place, we could improve the computational efficiency and the accuracy of the 3-D scene graph construction framework in the following research. The present framework interprets the image frames using deep learning for semantics and a pose estimation algorithm for physical attributes. Although the algorithms run in real time when used individually, each of them generally requires the entire computation capability that a single processing unit offers. We proposed KGE for the sake of efficiency, but a single processing unit cannot guarantee real-time operation. Furthermore, the accuracy of the framework depends on the performance of each module. A part of the modules achieves human-level performance, yet the remaining part needs further improvements. We supplemented the framework with human knowledge and the prior from other knowledge base by proposing SDR, but the framework still misses a few corner cases. In other words, enhancement in both deep-learning-based recognition modules and the pose estimation module would benefit the proposed 3-D scene graph construction framework.

For the third point, we could investigate the architecture of the proposed 3-D scene graph construction framework. We have built up a set of modules that take distinct roles. However, an end-to-end architecture or integration of a part of modules to form a bigger building block is possible. It is reported that the integration of submodules, whose functions are related, results in performance improvement when combined properly. This formulation, known as multitask learning [35], helps the integrated modules to generalize over the training data and to avoid overfitting. Candidates for the incorporation include the pair of relation extraction and SDR.

Last but not least, we could explore the application areas that might benefit from 3-D scene graph and expand the demonstrated applications presented in this paper. For the exploration of the application areas, environment description in natural language and semantic segmentation for 3-D spaces are representative. The environment description could support the blind to understand the surrounding spaces and avoid possible dangers. The semantic segmentation for 3-D spaces would aid 3-D dense modeling/reconstruction of 3-D spaces. For the expansion of the presented applications, the current VQA system can be upgraded to full-sentence VQA system or linked to an active search system. In addition, the task planning algorithm introduced in this paper only verified the feasibility of the 3-D scene graph's utilization in task planning. We could formulate the related research issues in a more delicate and mathematical way for the state-of-the-art level performance.

## IX. Conclusion

In this paper, we defined 3-D scene graph and proposed the 3-D scene graph construction framework. 3-D scene graph represents the surrounding environments in a sparse and semantic manner, providing intelligent agents with an effective environment model to store collected information and retrieve the information for reasoning and inference. The well-known graph structure of the 3-D scene graph offers intuitive usability and broad scalability. We verified the applicability and the accuracy of the 3-D scene graph through two major applications: 1) VQA and 2) task planning. Moreover, the proposed 3-D scene graph construction framework generates 3-D scene graphs representing environments upon receiving multiple observations in the form of image frames. The 3-D scene graph construction framework handles the input image frames both efficiently through the keyframe groups and effectively through the SDR algorithm. The experimental results established the four properties of an effective environment model for 3-D scene graph: 1) accuracy; 2) applicability; 3) usability; and 4) scalability.

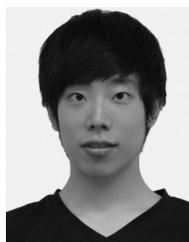

**Ue-Hwan Kim** received the B.S. and M.S. degrees in electrical engineering from the Korea Advanced Institute of Science and Technology, Daejeon, South Korea, in 2013 and 2015, respectively, where he is currently pursuing the Ph.D. degree.

His current research interests include service robot, cognitive IoT, computational memory systems, and learning algorithms.

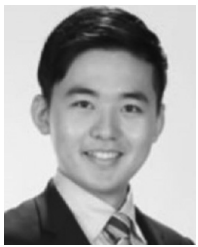

**Jin-Man Park** received the B.S. degree in electrical and electronic engineering from Yonsei University, Seoul, South Korea, in 2015, and the M.S. degree in the robotics program from the Korea Advanced Institute of Science and Technology, Daejeon, South Korea, in 2017, where he is currently pursuing the Ph.D. degree.

His current research interests include service robot, natural language processing, and visual question answering.

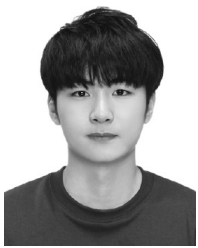

**Taek-Jin Song** received the B.S. degree in electrical engineering from the Korea Advanced Institute of Science and Technology, Daejeon, South Korea, in 2017, where he is currently pursuing the integrated master's and Doctoral degrees.

His current research interests include service robot, visual question answering, and learning algorithms.

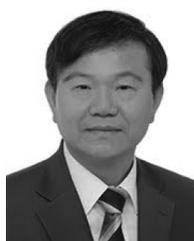

**Jong-Hwan Kim** (F'09) received the Ph.D. degree in electronics engineering from Seoul National University, Seoul, South Korea, in 1987.

Since 1988, he has been with the School of Electrical Engineering, Korea Advanced Institute of Science and Technology (KAIST), Daejeon, South Korea, where he is leading the Robot Intelligence Technology Laboratory as a KT Endowed Chair Professor. He is the Director for both KoYoung–KAIST AI Joint Research Center and Machine Intelligence and Robotics Multisponsored Research and Education Platform. He has authored 5 books and 5 edited books, 2 journal special issues, and around 400 refereed papers in technical journals and conference proceedings. His current research interests include intelligence technology, machine intelligence learning, and AI robots.